**Peri-AIIMS: <u>Peri</u>operative <u>A</u>rtificial <u>I</u>ntelligence Driven <u>I</u>ntegrated <u>M</u>odeling of <u>S</u>urgeries using Anesthetic, Physical and Cognitive Statuses for Predicting Hospital Outcomes**

Sabyasachi Bandyopadhyay, PhD[1,2]; Jiaqing Zhang, MS[2,3], Ronald L. Ison, MS[2,4]; David J. Libon, PhD[2,5]; Patrick Tighe, MD, MS[2,4]; Catherine Price, PhD[2,4,6,†,*]; Parisa Rashidi, PhD[1,2,*]

[1] J. Crayton Pruitt Family Department of Biomedical Engineering, University of Florida, Gainesville, FL, USA

[2] Perioperative Cognitive Anesthesia Network[SM], UF Health, University of Florida, Gainesville, FL, USA

[3] Department of Electrical and Computer Engineering, University of Florida, Gainesville, FL, USA

[4] Department of Anesthesiology, College of Medicine, University of Florida, Gainesville, FL, USA

[5] Department of Geriatrics and Gerontology, Department of Psychology, New Jersey Institute for Successful Aging, School of Osteopathic Medicine, Rowan University, Glassboro, NJ, USA

[6] Department of Clinical and Health Psychology, College of Public Health and Health Professions, University of Florida, Gainesville, FL, USA

*These authors contributed equally.

† Corresponding author: Catherine Price, PhD; Department of Clinical and Health Psychology, Gainesville, FL, 32610; Phone: 352-273-5272; Email: cep23@phhp.ufl.edu

**Financial Disclosures:** This work was conducted at the University of Florida. C.P. was supported by R01 NR014810 awarded by the National Institute of Nursing Research. C.P. was also supported by R01 NS082386 awarded by the National Institute of Neurological Disorders and Stroke. C.P. was supported by K07AG066813 by the National Institutes of Health. C.P. and P.T. were both supported by R01AG055337 by the National Institute on Aging, the National Center for Advancing Translational Science, rand the University of Florida. P.T. was supported by K07AG073468 by the National Institutes of Health. P.R. was supported by National Science Foundation CAREER award 1750192.

**Conflicts of Interest:** None.

**Word Counts:** Abstract: 301, Introduction: 597, Discussion: 1231, Overall: 5038.

**Running Head:** PERIOPERATIVE AI MODELING FOR HOSPITAL OUTCOMES

**Author Contributions:**

Sabyasachi Bandyopadhyay: This author helped with designing the study, preprocessing, analyzing, and interpreting the data, and drafting the manuscript.

Jiaqing Zhang: This author helped with processing the study, interpreting the data, and drafting and revising the manuscript.

Ronald L. Ison: This author helped with conceptualizing the study, and interpreting the data.

David J. Libon: This author helped with conceptualizing the study, interpreting the data, and revising the manuscript substantively.

Patrick Tighe: This author helped with conceptualizing the study and interpreting the data.

Catherine Price: This author helped with conceptualizing the study, acquiring and interpreting the data, and revising the manuscript substantively.

Parisa Rashidi: This author helped with designing the study, interpreting the data, and revising the manuscript substantively.

**ABSTRACT**

The association between preoperative cognitive status and surgical outcomes is a critical, yet scarcely explored area of research. Linking intraoperative data with postoperative outcomes is a promising and low-cost way of evaluating long-term impacts of surgical interventions. In this study, we evaluated how preoperative cognitive status as measured by the clock drawing test contributed to predicting length of hospital stay, hospital charges, average pain experienced during follow-up, and 1-year mortality over and above intraoperative variables, demographics, preoperative physical status and comorbidities. We expanded our analysis to 6 specific surgical groups where sufficient data was available for cross-validation. The clock drawing images were represented by 10 constructional features discovered by a semi-supervised deep learning algorithm, previously validated to differentiate between dementia and non-dementia patients. Different machine learning models were trained using 5-fold cross-validation to classify postoperative outcomes in hold-out test sets. The models were compared with respect to their relative performance, time complexity, and interpretability. Shapley Additive Explanations (SHAP) analysis was used to find the most predictive features for classifying different outcomes in different surgical contexts. Relative classification performances achieved by different feature sets showed that the perioperative cognitive dataset which included clock drawing features in addition to intraoperative variables, demographics, and comorbidities served as the best dataset for 12 of 18 possible surgery-outcome combinations. Interpretability analysis showed that duration of surgery was the most significant predictor of adverse outcomes, followed by the mean alveolar concentration of isoflurane or sevoflurane used during anesthesia. Disruptions in baseline correlations between intraoperative variables in different surgeries and the relative directionality of these variables in SHAP plots revealed that low average blood pressure and high standard deviation of blood pressure (ie, risk of blood loss during surgery) predicted adverse outcomes. Among the clock features, differences in clock size was the most significant predictor of adverse outcomes.

**Glossary of Terms**

**AdaBoost** = adaptive boosting; **ADI** = area deprivation index; **AI** = artificial intelligence; **ASA** = American Society of Anesthesiologists; **AUC** = area under the curve; **AUROC** = area under the receiver operating curve; **AVG NIBP** = average deviation of noninvasive blood pressure; **BMI** = body mass index; **CatBoost** = categorical boosting; **CD** = cognitive disorder; **CDT** = clock drawing test; **DB** = diabetes; **DL** = deep learning; **HCl** = hydrochloride; **HIPAA** = Health Insurance Portability and Accountability Act; **HL** = hyperlipidemia; **HT** = hypertension; **iso-sev-MAC** = mean alveolar concentration of isoflurane and sevoflurane; **LOS** = length of hospital stay; **LR** = logistic regression; **MD** = movement disorder; **ML** = machine learning; **MLP** = multilayered perceptron; **MMSE** = Mini Mental State Exam; **MRI** = magnetic resonance imaging; **MME** = morphine milligram equivalent; **Peri-AIIMS** = perioperative artificial intelligence–based integrated modeling of surgeries; **RF-VAE** = relevance-factor variational autoencoder; **SA** = sleep apnea; **SD NIBP** = standard deviation of noninvasive blood pressure; **SHAP** = Shapley Additive Explanations; **UF** = University of Florida; **VAE** = variational autoencoder; **XGBoost** = extreme gradient boosting

**INTRODUCTION**

The risk of postoperative complications in older adults depends on preoperative health modulated by the type of surgery and quality of anesthesia.[1] Subjective measures of comorbidity such as the American Society of Anesthesiologists (ASA) physical status score are the most commonly used metrics for assessing preoperative health risk.[2] However, such assessments may not be sufficient for predicting postoperative outcomes.[3,4] Recently, machine learning (ML) models and other heuristic scores have been used to stratify the risk of mortality and other specific outcomes using a combination of preoperative and intraoperative indices.[5–13] However, studies that explored the relationships between pre-anesthesia physical status and intraoperative variables were generally undertaken for a specific surgery, except for a few studies[8,13] where the risk of multiple adverse outcomes was stratified over a wide variety of surgery types. No study has yet investigated the impact of preoperative screeners beyond comorbidities for the prediction of postoperative outcomes.

Postoperative outcome prediction models built on intraoperative variables may be improved by examining the additional contribution of preoperative cognitive screeners beyond comorbidities. Preoperative cognitive screeners have been shown to predict the risk of postoperative delirium and mortality in older adults.[14] Memory and executive subitems from cognitive screeners such as the Mini Mental State Exam (MMSE)[15] have been shown to predict postoperative delirium after cardiac surgery[16] and screeners from the Mini-Cog[17] could predict outcomes after orthopedic surgery.[18] However, these studies did not examine the importance of preoperative cognitive screener performance in addition to that of intraoperative variables such as anesthesia type, anesthesia depth, and duration of surgery for predicting postoperative adverse outcomes. Without this exploration, the value of cognitive screeners cannot be fully assessed. This topic has become increasingly relevant due to the increasing number of older adults nationally,[19] combined with the facts that older adult patients represent at least 50% of all surgeries[20] and the expected rate of dementia is increasing on average 6% across the United States.[21]

For this investigation, we explored the additive effect of preoperative cognitive screening performance as measured by the command and copy clock drawing test (CDT) over and above intraoperative variables, preoperative physical status, demographics, and comorbidities on postoperative outcomes for older adults electing surgeries within a tertiary care medical center. The CDT is a simple yet intuitive measure of cognitive function. The test is comprised of 2 parts: (A) the *command* condition, where the subject is asked to "draw a clock, put in all the numbers, and set the hands to 10 after 11," followed by (B) the *copy* condition, where the subject is asked to copy a model clock. Accurate clock construction relies on a seamless and integrated operation of several cognitive abilities, and subtle changes in clock drawing can reveal underlying cognitive deficits.[22] For example, clock drawing to command condition requires the simultaneous syntactic comprehension of verbal instructions, recalling the semantic attributes of a clock, working memory, accurate mental planning, and fine motor skills.[23] Similarly, in the copy condition, clock construction is reliant on visual scanning abilities, visuoconstruction, and executive functioning.[24] We represented clock drawings using a semi-supervised deep learning (DL)–based 10-dimensional representation that was previously validated to capture unique constructional features of the CDT relevant to dementia.[25] We hypothesized that irrespective of the surgical context, preoperative cognition as encoded by this DL representation of clock drawings would inform classification of postoperative outcomes that are explicitly unrelated to cognition. Figure 1 highlights the process flow in the perioperative surgical environment network and shows how in this study we integrated different data modalities arising during different stages of this process into an analytic framework for predicting postoperative adverse outcomes.

## METHODS

Participants

Data were collected via a federally funded investigation at the University of Florida (UF) and UF Health Shands hospital. The UF Institutional Review Board approved the study. Health Insurance Portability and Accountability Act (HIPAA) waiver and data were extracted from the medical records through an honest data broker. The study was carried out in accordance with the guidelines set by the Declaration of Helsinki, respective university guidelines, and TRIPOD criteria.[26]

The original surgical dataset was comprised of 22,473 patients aged 65 or older who completed an in-person preoperative assessment for elective surgeries between January 2018 and December 2019. Of this data, we possessed complete intraoperative data for 18,037 patients, preoperative physical status information for 12,422 patients, complete demographics (age, race, sex, ethnicity, education, and area deprivation index [ADI][27–29]) for 10,055 patients, and clock drawings for 11,777 patients. Supplemental Figure 1 shows an integrated consensus diagram illustrating the inclusion of different data modalities used for creating the datasets reported in this study. As demonstrated in Supplemental Figure 1, stepwise combination of these data-modalities was carried out to develop the 3 datasets primarily used for our analysis. These 3 primary datasets were divided according to surgery types (Supplemental Figure 1) to create surgery-specific intraoperative, perioperative, and perioperative cognition datasets used for analysis.

Categories of Variables

The primary variables used in this study can be divided into 4 categories:

- *Intraoperative variables*: type of surgery, duration of surgery (in minutes), propofol dose (mg), oral morphine milligram equivalent (MME) dose (mg), anesthetic depth as represented by the mean alveolar concentration of isoflurane or sevoflurane (iso-sev-MAC), mean and standard deviation of noninvasive blood pressure (SD NIBP), and doses of the vasopressors phenylephrine (mcg) and ephedrine (mg).

- *Demographics*: age, sex, race, ethnicity, years of education, and ADI (as calculated based on the patient's location of residence at the time of surgery). For analysis consisting of all surgeries, the surgery type was one-hot encoded. Surgery type was defined as the primary surgical service operating on the patient. For analysis with individual surgeries, surgery type was not a viable feature.

- *Preoperative physical status*: ASA physical status score,[30] body mass index (BMI) at time of admission, frailty score,[31] and comorbidities.

Admission BMI was not used in the final analyses because greater than 60% of the data pertaining to this variable were missing. Comorbidities present in our dataset included sleep apnea, diabetes, hyperlipidemia, hypertension, movement disorders, and cognitive disorders.

- *Preoperative cognitive screener*: in this study, only the CDT was used as a preoperative cognitive screener variable. Every participant completed a preoperative clock drawing to command and copy condition as part of a hospital wide preoperative screening program for older adults.[32] Per published instructions, individuals were required to "draw the face of a clock, put in the numbers, and set the hands to ten after eleven," and then copy a model clock.[33] Clock drawings were captured using digital pen technology.[34,35] Only the final output image of the digital clock drawing was used for this analysis. Final output images of clock drawings to command and copy conditions were projected on a 10D DL latent space described in our previous work.[25,36] The deep CDT representation used in this study originated from a latent representation developed using semi-supervised training of a relevance-factor variational autoencoder (RF-VAE) network to classify dementia from non-dementia participants.[25] Briefly, a RF-VAE network was trained using 23,521 unlabeled clock drawings collected as part of a routine check-up in the preoperative setting.[32] The RF-VAE discovered 10 unique constructional aspects of the CDT that were validated to be different in people with dementia versus healthy peers using a smaller classification dataset. This study excluded participants if they were non-fluent in English, had <4 years of education or had some visual, motor or hearing impairment that severely limited their ability to execute a CDT. A prior publication reported a simpler version of this representation, which studied the same constructional features of clock drawings in the form of a 2-dimensional representation, and assessed its ability to classify between dementia and non-dementia participants belonging to the same classification dataset.[36] These datasets have been described in previous studies.[22,34,37–39] Participants whose clock drawings were used to train the RF-VAE network overlapped with those in our current study. The 10 unique constructional features described in the above study are shown in Supplemental Figure 2A-B. The relative occurrence in dementia samples described above is shown in Supplemental Figure 2C. All numerical variables were scaled to their respective maximum values to create a normalized dataset and all categorical variables were one-hot encoded. Six individual surgeries contained sufficient data for downstream analyses: orthopedics, neurosurgeries, cardiac and vascular surgeries, urologic surgeries, gynecologic surgeries, and otolaryngologic surgeries.

Outcomes

In this study, postoperative outcomes of interest were (A) *length of hospital stay* (LOS; hours), (B) *hospital charges* (dollars), (C) *1-year mortality*, and (D) *average pain* during follow-up period. Mortality classification was not performed for individual surgeries due to the low number of occurred deaths falling in each fold of cross-validation and in the holdout test dataset. Outcomes were binarized in the following manner:

- LOS = 0: same day discharge, 1: longer than one day,

- Hospital charges = 0: less than $30,000, 1: greater than or equal to $30,000

- Mortality = 0: survived until one year after discharge, 1: died within one year of being discharged

- Average pain = 0: average pain = 0, 1: average pain ≥ 1. The average pain rating was calculated by averaging the pain score reported by patients according to the Brief Pain Inventory Short Form.[40] It was reported on a scale of 0 to 10. The same thresholds were used for individual surgeries.

<u>Cohorts</u>

Three different feature combinations were used to evaluate the potential additive effect of different data modalities to the classification performance: (A) the *intraoperative dataset*, created from anesthetic, blood pressure, and surgical features, (B) the *perioperative dataset*, developed by combining the *intraoperative dataset* with demographics, preoperative physical status, and comorbidities, and (C) the *perioperative cognitive dataset*, developed by combining the *perioperative dataset* with DL representation of clock drawings (Figure 2; Supplemental Figure 1). Command clock drawings and copy clock drawings were tested separately and in concert. The best performance amongst these 3 datasets was reported for each classification task. Table 1 shows the demographic characteristics and comorbidities for the patients included in this study.

<u>Procedure</u>

*Statistical Analysis*

For every surgery type, Pearson's product moment correlation matrices were calculated between the different intraoperative variables. Cohen's d effect sizes were used to evaluate the level of significance of the correlation values.[41]

*Artificial Intelligence*

An array of ML models including logistic regression (LR),[42] random forest,[43] naïve Bayes classifier,[44] extreme gradient boosting (XGBoost),[45] adaptive boosting (AdaBoost),[46] categorical boosting (CatBoost),[47] and a multilayered perceptron (MLP)[48,49] were trained on the training subset of each dataset in different surgical contexts. In each case, the hyperparameters of these models were optimized inside a 5-fold cross-validation setting. Area under the receiver operating curve (AUROC), accuracy, F1 score, sensitivity, specificity, and average precision were reported on the test-dataset. The best model was selected based on the best performance, least time complexity, and ease of interpretability. Shapley Additive Explanations (SHAP) analysis was performed for the best model. The test-dataset was bootstrapped 100 times with replacement to generate 95% confidence intervals.

**RESULTS**

Participants

Table 1 shows the demographic characteristics and comorbidities for the patients included in this study. Individuals in the study cohort were mostly White, college-educated, and aged greater than 65 years.

Statistical Analyses

We evaluated the cross-correlation between the various anesthetic and surgical features (Figure 3). Correlation values were categorized as high if they were greater than 0.5, moderate if they were between 0.3 and 0.5, and low if they were between 0.1 and 0.3 according to Cohen's d effect size criteria.[41] Duration of surgery had a high positive correlation with oral MME in cardiovascular (Figure 3D), urologic (Figure 3E), and gynecologic (Figure 3F) surgeries, as well as when considering all surgeries together (Figure 3A). In orthopedic (Figure 3B), neurologic (Figure 3C), and otolaryngologic (Figure 3G) surgeries, surgery duration had a moderate positive correlation with oral MME. Phenylephrine had a high positive correlation with average blood pressure and standard deviation of blood pressure in all cohorts (Figure 3). Average blood pressure had a moderate positive correlation with standard deviation of blood pressure in orthopedic (Figure 3B), neurologic (Figure 3C), urologic (Figure 3E), gynecologic (Figure 3F), and otolaryngologic (Figure 3G) surgeries. In cardiovascular surgeries (Figure 3D) and when all surgeries were considered together (Figure 3A) this correlation became low. Ephedrine had a moderate negative correlation with average blood pressure in gynecologic (Figure 3F) and otolaryngologic (Figure 3G) surgeries. Additionally, iso-sev-MAC had a moderate negative correlation with propofol in orthopedic (Figure 3B), urologic (Figure 3E), gynecologic (Figure 3F), and otolaryngologic (Figure 3G) surgeries.

These relations are instrumental to interpreting the surgical characteristics of patients who experienced adverse outcomes.

Artificial Intelligence

*Model Performances*

The best ML models for each outcome and each surgical context are reported in Table 2, as is the best dataset used for predicting hospital outcomes and its corresponding division of samples between classes 0 and 1. Table 3 highlights which dataset was deemed most predictive in each case. In 12 of 22 cases, the *Perioperative cognitive dataset* generated the best model, in 5 instances the *Perioperative dataset* gave the best model, and in 5 other instances, the *Intraoperative dataset* gave the best model. Charges classification was performed superlatively across all surgeries, and LOS classification was performed moderately in orthopedics, urology and gynecology surgeries. However, average pain during follow-up proved to be a difficult classification problem. One-year mortality, which could only be classified for all surgeries, had very low precision and F1 score indicating the prevalence of many false positives compared to true positives. This is partially due to the heavy class imbalance present in this dataset.

*Interpretability analysis*

SHAP analysis was performed on each best model to find the top 10 most informative features. Figure 4 shows the SHAP results for the different surgery groups.

In 16 of 18 SHAP plots, surgery duration was the most important predictor. In the other 2 plots (Figures 3C, 3I) it was the second most important predictor. Besides surgery duration, iso-sev-MAC appears as one of the top 2 most predictive features in 7 of 18 situations. The baseline correlations present between the surgical variables in each surgery were shown in Figure 2. The Results section detailed which of these associations were maintained and which were reversed in the SHAP plots. This indicates which correlations were broken in patients who experienced adverse outcomes (ie, longer LOS, higher charges and more average pain).

<u>Intraoperative Characteristics of Adverse Outcomes</u>

*Orthopedic Surgery*

Figure 4A reveals that a higher LOS was associated with high values of SD NIBP and low values of average deviation of noninvasive blood pressure (AVG NIBP) during surgery. This is opposite to the correlation expected from Figure 3B. The same pattern is further exaggerated in average pain where AVG NIBP during surgery is the most important predictor, and SD NIBP is the third most important predictor of greater pain (Figure 4C).

*Neurosurgery*

Figure 4D shows that the model associated higher doses of ephedrine sulphate, high AVG NIBP, and high SD NIBP during surgery with longer LOS. This pattern is opposite to the correlations found between these variables in Figure 3C. The same pattern is repeated in average pain classification (Figure 4F); additionally, ephedrine sulphate dose is a more important predictor than the blood pressure values. Figure 4E reveals that longer durations of surgery, lower values of iso-sev-MAC (topmost features), and lower AVG NIBP but higher SD NIBP (less important) were associated with higher hospital charges. These patterns are also in contrast with the correlations to be expected in the neurosurgery group from Figure 3C.

*Cardiac and Vascular Surgery*

Figure 4I shows that in cardiovascular surgeries patients who experienced a low SD NIBP while been exposed to high doses of phenylephrine hydrochloride (HCl) pressor were predicted to experience greater average pain. Figure 3D shows that cardiovascular surgery patients should have high SD NIBP if they received high volumes of phenylephrine. In comparison to the orthopedic and neurologic surgeries, LOS and charges in this particular surgery group were more amply predicted by preexisting comorbidities.

*Urologic Surgery*

Figures 4J through Figure 4L show no deviations from the baseline correlation patterns found in Figure 3E.

*Gynecologic Surgery*

Figure 4N shows that hospital charges in gynecologic surgeries are solely dependent on and driven by the surgery duration. Figure 3F shows that in the overall gynecologic surgical group ephedrine sulphate is negatively correlated with phenylephrine HCl and AVG NIBP. In comparison, Figure 4O shows that higher postoperative pain in this cohort was predicted by low phenylephrine HCl, low ephedrine sulphate, high SD NIBP (more important), and high AVG NIBP.

*Otolaryngologic Surgery*

Figure 4P shows that people who experienced high SD NIBP while simultaneously showing low mean intraoperative blood pressure were predicted to have longer LOS. This is the opposite of the general positive correlation pattern expressed between these 2 variables in this surgical group (Figure 3G). The same pattern is exaggerated in Figure 4R (average pain). Additionally, Figure 4R shows that the incidence of higher iso-sev-MAC together with higher propofol, two negatively correlated variables predicted greater average pain during follow-up. Figure 4Q shows that in this group, charges are primarily driven by the duration of surgery.

The loss of a positive association between AVG NIBP and SD NIBP and specifically higher SD NIBP with lower AVG NIBP predicted prolonged LOS and/or greater average pain in 3 of 6 surgical contexts (orthopedic, cardiovascular, and otolaryngological).

Comorbidities and Disparities in Adverse Outcomes

*Orthopedic Surgery*

Figure 4A shows that being older and more frail was associated with longer LOS after orthopedic surgery. Figure 4B shows that patients with higher education were predicted to pay less hospital charges. Figure 4C shows that being older was associated with reporting less average pain. Also, higher frailty and hypertension were associated by the best model with higher average pain.

*Neurosurgery*

Figure 4E shows that in neurosurgery, White patients were predicted to pay less hospital charges than non-White patients, while patients with hyperlipidemia and movement disorders were predicted to pay higher charges than others.

*Cardiovascular surgery*

Figure 4G shows that hyperlipidemia, hypertension, and sleep apnea predicted longer LOS, while diabetes predicted lower LOS. Being biologically female and/or Black was associated by the model with lower LOS. Figure 4H shows that higher education, hyperlipidemia, and hypertension predicted for higher charges.

*Urologic Surgery*

Figure 4J shows that higher education predicted lower LOS; also, hyperlipidemia and diabetes predicted lower LOS in urologic surgeries. Diabetes also predicted lower charges for this surgery group (Figure 4K). Figure 4L shows that older age predicted lower self-reported pain in this surgery group.

*Gynecologic Surgery*

Figure 4M shows that higher education and higher ASA scores predicted greater LOS. Hyperlipidemia predicted higher LOS, but hypertension predicted lower LOS in this group. Figure 4O shows that older age predicted lower reported pain, and higher ADI predicted greater pain in gynecologic surgeries.

*Otolaryngologic Surgery*

Adverse outcomes in otolaryngologic surgeries were not strongly predicted by any comorbidities or demographic disparities.

Older people were predicted to report less pain in 3 surgical contexts: orthopedic, urologic, and gynecologic. Hyperlipidemia predicted longer LOS in 2 surgical contexts: gynecologic and cardiovascular. Hyperlipidemia also predicted higher hospital charges in 2 surgeries: neurologic and cardiovascular.

Clock Drawing Characteristics in Adverse Outcomes

Preoperative clock drawing characteristics served as important predictors in 9 of 18 surgery-outcome combinations (Table 3). Among these, clock size was the most important cognitive predictor in 5 surgery-outcome contexts: 2 orthopedic (LOS, charges), 1 neurosurgery (charges), 1 cardiovascular (charges), 1 gynecology (LOS) (Figures 4A, 4B, 4E, 4H, 4M). In *orthopedic surgery,* larger clock size predicted greater LOS and greater hospital charges (Figures 4A, 4B). In contrast, in *neurosurgery, cardiovascular surgery,* and *gynecologic surgery* a larger clock predicted lower charges and lower LOS, respectively (Figures 4E, 4H, 4M). Clock size was also an important predictor in 2 of the remaining 4 surgery-outcome combinations (urology-charges, and otolaryngology-LOS) (Figures 4K, 4P). Therefore, clock size was found to be the most prevalent predictive cognitive characteristic in this report. Other frequently important clock features were rotated ellipse (normal or vertical), upward displaced hands, ovate-obovate shape, and obtuse angle between hands. Rotated ellipse and/or rotated vertical ellipse predicted longer LOS in *orthopedic surgery* (Figure 4A)*,* lower hospital charges in *neurosurgery* (Figure 4E), lower hospital charges in *cardiovascular surgery* (Figure 4H), higher pain in *urologic surgery* (Figure 4L), and lower LOS in *gynecologic surgery* (Figure 4M). It was also an important predictor in otolaryngology-LOS, but the directionality could not be easily understood (Figure 4P). Therefore, rotated elliptical preoperative clocks were the second most prevalent cognitive predictor in our study. Upward displaced clock hands predicted lower charges in *orthopedic surgery* (Figure 4B), lower LOS, higher charges, and higher pain in *urologic surgery* (Figures 4J, K), and higher LOS in *otolaryngology surgery* (Figure 4P). Obovate clocks predicted higher charges in *orthopedic surgery* (Figure 4B)*,* higher LOS and lower pain in *urologic* *surgery* (Figures 4J, 4L)*,* and lower LOS in *gynecology surgery* (Figure 4M). An obtuse angle between hands predicted lower charges in *orthopedic surgery* (Figure 4B), *cardiovascular surgery* (Figure 4H), and higher LOS in *urologic surgery* (Figure 4J). Therefore, after clock size, 2 clock shape features (elliptical and obovate) and 2 hand placement features (upward displacement and wide angle between) were the most significant features.

**DISCUSSION**

This study shows that a complex array of perioperative network cognitive, anesthetic, and physical variables predicted postoperative adverse outcomes such as longer length of stay, higher hospital charges, higher average pain, and mortality. Six different surgery types including *orthopedic*, *neurologic*, *cardiovascular*, *urologic*, *gynecologic,* and *otolaryngologic* surgeries were individually studied. Machine learning classifiers were developed in each case to maximize recognition of adverse outcomes and these models were interpreted to understand the degree and directionality of influence shown by the perioperative variables. This comprised an integrated AI-driven modeling of surgeries using perioperative information including cognition.

Intraoperative Variable Networks

A study of the cross-correlation patterns of the intraoperative variables suggested significant positive correlation between duration of surgery, OME, and iso-sev-MAC, and a secondary group of strong positive associations between AVG NIBP, SD NIBP, and phenylephrine. Phenylephrine HCl and ephedrine sulphate, both vasopressors, were negatively correlated to each other. The first group of strong positive correlations represents the *anesthetic and analgesic network,* while the second group of positive correlations represents the *hemodynamic management network.* These represent the 2 primary housekeeping tasks during any surgical procedure. A graphical illustration of these associations is shown in Figure 5.

Additionally, certain surgical contexts, such as orthopedic, urologic, gynecologic, and otolaryngologic surgeries exhibited a strong negative association between propofol and iso-sev-MAC. Our findings indicated that in different surgery-outcome combinations these general correlation patterns were disrupted in patients suffering from long-term adverse outcomes. This breakdown of correlation patterns was found in the relative "directionality of influence" of these features while being utilized by the ML models to predict adverse outcomes. The *hemodynamic network* demonstrated higher disruption in adverse outcomes than the *anesthetic and analgesic network.* The *hemodynamic network* showed a deviation from expected correlation 9 of 12 times while the *anesthetic network* did so 2 of 12 times when their features were significant predictors of adverse outcomes. This observation combined with the fact that the most common disruption in the *hemodynamic network* was higher variance in blood pressure with low AVG NIBP indicates that blood loss during surgery was a driver of postoperative adverse outcomes. Despite not being disrupted as much as the *hemodynamic network,* the *anesthetic and analgesic network* was the more important of the two in predicting adverse outcomes because duration of surgery was the most significant predictor in our study. Furthermore, iso-sev-MAC was within the top 2 most important predictors in 7 of 18 scenarios, while oral MME was within the top 10 predictors in 7 of 18 scenarios, suggesting that a combination of longer surgery and greater depth of anesthesia and analgesia concertedly generated adverse postoperative effects.

Significance of Clock Drawing Features

Preoperative cognition was represented by the CDT. Specifically, the CDT to command and copy condition was projected onto a pretrained RF-VAE, DL encoder which represented the clock drawing as a combination of 10 unique, mutually disentangled constructional features of the clockface.[25] These features had been previously validated to distinguish between clocks drawn by dementia and non-dementia patients. Among the clock drawing features that were investigated, clock size was the most important predictor. This was followed by distorted clock shape represented by rotated ellipticity and obovateness, as well as abnormal time representation as seen by the upward displacement and obtuse angle between the clock hands. Successful clock drawing requires the coordination of multiple neurocognitive operations including access to semantic information, mental planning, visuospatial reasoning, and motor ability.[24] In a community-dwelling, non-dementia sample, clock drawing has been linked to the integrity of magnetic resonance imaging (MRI) white matter connexions[50] and a large array of neuropsychological test parameters,[51] further illustrating that multiple neurocognitive operations must be coordinated for successful test performance. Aberrations in clock size and shape are frequently present in patients with greater dysexecutive impairment.[33,52,53] Problems in the correct representation of time (ie, incorrectly drawing the clock hands) has been shown to the associated with a combination of semantic and dysexecutive deficits.[24] Thus, the digital clock drawing test is parsimonious, provides data not otherwise available using other tests, and is a well-tolerated means by which to assess perisurgical cognitive abilities.

Relative Importance of Different Variable Groups

Although the perioperative cognitive variables demonstrated strongest classification performance in 12 of 18 surgery-outcome combinations, the difference in performance from the other variable sets, especially the intraoperative variables, were not considerable (Supplemental Figure 3). This indicated that intraoperative variables were the strongest predictors of postsurgical complications. However, in many instances the addition of clock features either improved the specificity or improved sensitivity-specificity balance; for example, a) in all surgeries - LOS, charges, mortality (Supplemental Table 1), b) charges in *orthopedic* surgery (Supplemental Table 2), c) LOS and charges in *neurosurgery* (Supplemental Table 3), d) LOS in *urologic* surgery (Supplemental Table 5), e) LOS in *gynecologic* surgery (Supplemental Table 6). This indicated that preoperative cognition had an impact on postoperative recovery in multiple surgical contexts. These observations will be explored in further detail in a future study.

Strengths

Our study has unique strengths. First, the data reported above explored a wide array of preoperative and intraoperative variables for retrospectively classifying postoperative outcomes using ML. This study is the first of its kind to perform an integrated assessment of different preoperative datasets including cognitive features for predicting postsurgical outcomes. Machine learning classifiers with different inductive biases were used to improve the generality of classification performance over the different datasets for the diverse surgery-outcome combinations. Application of ML models allowed the modeling of nested, nonlinear relationships within the multimodal datasets. Interpretability analysis using SHAP helped us understand the characteristics of adverse outcomes from the lenses of intraoperative, preoperative physical status, and cognitive variables. This led to better understanding of the perioperative stressors leading to postoperative adverse outcomes. Others posit novel, provocative hypotheses which should be explored in future studies. Such explorations can help reduce healthcare costs, deliberate better prognosis, reduce the rate of readmissions, and significantly alter health management beyond the hospital setting.

Limitations

This study has certain limitations. The size of the initial dataset was significantly impacted by missing entries. The final dataset used for classification was one-fourth of the size of the initial dataset. Our study was limited by the binary classification setting. However, the use of binary classification allowed us to test diverse surgeries within the same framework. In the future, individual surgeries should be studied with appropriate postoperative outcomes, and DL-based clustering or self-supervised learning can be used to learn compressed representations of the surgeries themselves. These ideas combined with multitask learning can generate a deep surgical signature helpful for downstream classification of postoperative complications. The use of transfer learning or multitask learning can create information that can be translated to other datasets/tasks. We will also explore the hypothesis of the positive correlation between cognitive well-being and postoperative recovery in future works.

Conclusion

In summary, this work represents the first of its kind to perform an integrated AI-driven modeling of a diverse group of surgeries for classifying multiple postoperative outcomes. We show that the perioperative network of variables including preoperative cognitive factors can provide additional information for predicting postoperative outcomes over and beyond intraoperative variables. However, it also emphasizes the importance of intraoperative anesthetic, analgesic, and hemodynamic factors in this regard. This report illustrates that hemodynamic mismanagement and increased surgery duration, anesthesia depth, and analgesia dosage all contributed significantly to development of postoperative adverse effects. It also indicated that preoperative cognitive ability might have a protective effect on postsurgical health.

**ACKNOWLEDGEMENTS**

We would like to acknowledge Shawna Amini for managing the datasets used for this project.

The content is solely the responsibility of the authors and does not necessarily represent the official views of the National Institute on Aging, National Institute of Nursing Research, National Institute of Neurological Disorders and Stroke, National Institutes of Health, National Center for Advancing Translational Science, or University of Florida.

**TABLES**

**Table 1.** - Patient Demographics and Comorbidities

| | All Surgeries | Orthopedics | Neurology | Cardiac and Vascular | Urology | Gynecology | Otolaryngology |
|---|---|---|---|---|---|---|---|
| N | 6221 | 916 | 542 | 461 | 639 | 203 | 395 |
| Age (years) | 73.3 ± 6.0 | 72.0 ± 5.6 | 73.0 ± 5.3 | 73.9 ± 6.1 | 74.1 ± 6.2 | 72.5 ± 5.7 | 73.9 ± 6.3 |
| Sex (Male, Female) (%) | 50.3, 49.7 | 44.3, 55.7 | 55.1, 44.8 | 49.9, 50.1 | 71.0, 28.9 | 0.0, 100.0 | 54.9, 45.1 |
| Race (White, Black) (%) | 87.0, 7.3 | 86.6, 6.0 | 93.3, 2.7 | 86.9, 8.2 | 87.5, 6.7 | 88.7, 7.4 | 88.3, 5.3 |
| Ethnicity (non-Hispanic, Hispanic) (%) | 95.7, 2.0 | 93.8, 1.5 | 97.2, 2.0 | 94.3, 2.8 | 97.3, 2.0 | 98.0, 1.9 | 96.2, 1.8 |
| ADI | 60.2 ± 23.0 | 59.1 ± 23.0 | 54.7 ± 24.1 | 60.9 ± 21.9 | 61.0 ± 22.7 | 62.5 ± 21.9 | 58.7 ± 22.3 |
| Education (years) | 13.9 ± 2.9 | 14.4 ± 3.1 | 14.3 ± 2.8 | 13.4 ± 2.7 | 14.2 ± 3.1 | 13.8 ± 2.4 | 13.8 ± 2.8 |
| ASA (%) | 1: 0.1, 2: 11.5, 3: 79.5, 4: 8.9, 5: 0.01 | 1: 0.0, 2: 17.5, 3: 80.2, 4: 2.3, 5: 0.0 | 1: 0.4, 2: 10.7, 3: 86.7, 4: 2.2, 5: 0.0 | 1: 0.0, 2: 1.5, 3: 70.5, 4: 28.0, 5: 0.0 | 1: 0.0, 2: 13.1, 3: 84.2, 4: 2.7, 5: 0.0 | 1: 0.5, 2: 25.1, 3: 71.9, 4: 2.5, 5: 0.0 | 1: 0.0, 2: 14.9, 3: 80.0, 4: 5.1, 5: 0.0 |
| Frailty | 1.2 ± 1.3 | 1.4 ± 1.3 | 1.2 ± 1.4 | 1.3 ± 1.4 | 1.0 ± 1.3 | 1.1 ± 1.2 | 1.0 ± 1.3 |
| Comorbidities (%) | SA: 16.0, DB: 29.4, HL: 54.4, HT: 55.1, MD: 5.8, CD: 0.9 | SA: 16.5, DB: 23.8, HL: 57.2, HT: 59.9, MD: 4.6, CD: 0.9 | SA: 15.5, DB: 25.6, HL: 52.0, HT: 55.1, MD: 27.5, CD: 1.1 | SA: 12.1, DB: 33.8, HL: 64.4, HT: 41.4, MD: 2.1, CD: 0.4 | SA: 13.8, DB: 28.5, HL: 43.9, HT: 54.3, MD: 3.7, CD: 0.9 | SA: 8.4, DB: 20.7, HL: 43.3, HT: 52.7, MD: 1.9, CD: 0.5 | SA: 16.2, DB: 29.6, HL: 40.7, HT: 57.9, MD: 3.0, CD: 1.0 |

Abbreviations: ADI, area deprivation index; ASA, American Society of Anesthesiologists' Score; CD, cognitive disorder; DB, diabetes; HL, hyperlipidemia; HT, hypertension; MD, movement disorder; SA, sleep apnea.

**Table 2.** - Performance of Machine Learning Models on Long-Term Postsurgical

Outcomes Over Different Surgeries

| Surgery Type | Outcome | Best Dataset ($N_0$, $N_1$) | Best Model | AUC (95% CI) | Accuracy (95% C.I) | F1 Score (95% CI) | Precision (95% CI) | Sensitivity (95% CI) | Specificity (95% CI) |
|---|---|---|---|---|---|---|---|---|---|
| **All surgeries** | LOS | Peri-Op Cognitive (0: 3053, 1: 1908) | XGBoost | 0.93 (0.92 - 0.95) | 0.87 (0.85 - 0.89) | 0.83 (0.81 - 0.85) | 0.82 (0.79 - 0.84) | 0.85 (0.82 - 0.88) | 0.88 (0.86 - 0.90) |
| | Charges | Peri-Op Cognitive (0: 2341, 1: 2605) | XGBoost | 0.98 (0.97 - 0.99) | 0.93 (0.92 - 0.94) | 0.93 (0.92 - 0.95) | 0.94 (0.93 - 0.96) | 0.92 (0.91 - 0.94) | 0.94 (0.92 - 0.95) |
| | Average Pain | Peri-Op Cognitive (0: 2309, 1: 3084) | LR | 0.82 (0.80 - 0.84) | 0.77 (0.75 - 0.79) | 0.80 (0.78 - 0.82) | 0.80 (0.77 - 0.83) | 0.80 (0.77 - 0.82) | 0.73 (0.70 - 0.76) |
| | 1-year Mortality | Peri-Op Cognitive (0: 4740, 1: 226) | LR | 0.71 (0.66 - 0.77) | 0.70 (0.68 - 0.73) | 0.16 (0.11 - 0.20) | 0.09 (0.06 - 0.12) | 0.61 (0.50 - 0.73) | 0.71 (0.69 - 0.73) |
| **Orthopedics** | LOS | Peri-Op Cognitive (0: 207, 1: 662) | XGBoost | 0.84 (0.79 - 0.88) | 0.83 (0.79 - 0.88) | 0.90 (0.87 - 0.93) | 0.85 (0.82 - 0.89) | 0.95 (0.92 - 0.97) | 0.47 (0.35 - 0.58) |
| | Charges | Peri-Op Cognitive (0: 82, 1: 787) | LR | 0.99 (0.99 - 1.00) | 0.95 (0.92 - 0.97) | 0.97 (0.96 - 0.98) | 1.00 (1.00 - 1.00) | 0.95 (0.91 - 0.97) | 1.00 (1.00 - 1.00) |
| | Average Pain | Peri-Op (0: 156, 1: 752) | XGBoost | 0.69 (0.62 - 0.77) | 0.84 (0.80 - 0.88) | 0.91 (0.88 - 0.93) | 0.86 (0.82 - 0.89) | 0.97 (0.95 - 0.98) | 0.26 (0.16 - 0.38) |
| **Neurology** | LOS | Intra-Op (0: 430, 1: 693) | XGBoost | 0.92 (0.88 - 0.94) | 0.86 (0.82 - 0.90) | 0.89 (0.86 - 0.92) | 0.87 (0.83 - 0.91) | 0.91 (0.87 - 0.94) | 0.79 (0.72 - 0.84) |
| | Charges | Peri-Op Cognitive (0: 90, 1: 411) | LR | 0.93 (0.87 - 0.97) | 0.86 (0.80 - 0.91) | 0.91 (0.87 - 0.94) | 0.94 (0.89 - 0.98) | 0.88 (0.82 - 0.92) | 0.76 (0.60 - 0.91) |
| | Average Pain | Intra-Op (0: 328, 1: 795) | XGBoost | 0.74 (0.68 - 0.80) | 0.78 (0.74 - 0.82) | 0.86 (0.83 - 0.88) | 0.81 (0.77 - 0.86) | 0.90 (0.87 - 0.93) | 0.49 (0.41 - 0.58) |
| **Cardiac and vascular** | LOS | Peri-Op (0:192, 1: 260) | LR | 0.89 (0.83 - 0.93) | 0.83 (0.78 - 0.88) | 0.85 (0.80 - 0.90) | 0.86 (0.79 - 0.94) | 0.84 (0.76 - 0.92) | 0.82 (0.73 - 0.91) |
| | Charges | Peri-Op Cognitive (0: 93, 1: 359) | LR | 0.96 (0.93 - 0.98) | 0.91 (0.85 - 0.95) | 0.94 (0.90 - 0.97) | 0.98 (0.95 - 1.00) | 0.90 (0.82 - 0.95) | 0.95 (0.82 - 1.00) |
| | Average Pain | Intra-Op (0: 307, 1: 521) | LR | 0.70 (0.64 - 0.76) | 0.64 (0.59 - 0.70) | 0.69 (0.64 - 0.75) | 0.76 (0.70 - 0.83) | 0.63 (0.55 - 0.70) | 0.66 (0.56 - 0.74) |
| **Urology** | LOS | Peri-Op Cognitive (0: 383, 1: 148) | LR | 0.80 (0.71 - 0.87) | 0.77 (0.71 - 0.82) | 0.63 (0.52 - 0.72) | 0.57 (0.44 - 0.69) | 0.70 (0.59 - 0.81) | 0.80 (0.74 - 0.86) |
| | Charges | Peri-Op Cognitive (0: 295, 1: 236) | XGBoost | 0.96 (0.93 - 0.99) | 0.93 (0.89 - 0.96) | 0.91 (0.86 - 0.95) | 0.94 (0.88 − 1.00) | 0.88 (0.81 - 0.94) | 0.96 (0.92 − 1.00) |
| | Average Pain | Peri-Op Cognitive (0: 294, 1: 237) | LR | 0.75 (0.69 - 0.81) | 0.70 (0.64 - 0.76) | 0.70 (0.64 - 0.76) | 0.78 (0.68 - 0.86) | 0.64 (0.55 - 0.72) | 0.78 (0.69 - 0.88) |
| **Gynecology** | LOS | Peri-Op Cognitive (0: 123, 1: 50) | LR | 0.80 (0.67 - 0.89) | 0.72 (0.62 - 0.83) | 0.56 (0.33 − 0.71) | 0.52 (0.29 − 0.72) | 0.60 (0.35 - 0.81) | 0.78 (0.64 - 0.89) |
| | Charges | Intra-Op (0: 254, 1: 128) | LR | 0.99 (0.97 - 1.00) | 0.94 (0.89 - 0.97) | 0.95 (0.91 - 0.98) | 1.00 (1.00 - 1.00) | 0.91 (0.84 - 0.96) | 1.00 (1.00 - 1.00) |
| | Average Pain | Peri-Op (0: 167, 1: 33) | XGBoost | 0.74 (0.62 - 0.88) | 0.79 (0.70 - 0.88) | 0.87 (0.82 - 0.93) | 0.87 (0.80 - 0.95) | 0.88 (0.79 - 0.94) | 0.36 (0.12 - 0.68) |
| **Otolaryngology** | LOS | Peri-Op Cognitive (0: 228, 1: 111) | XGBoost | 0.88 (0.80 - 0.94) | 0.82 (0.75 - 0.88) | 0.69 (0.52 - 0.78) | 0.86 (0.71 - 0.95) | 0.58 (0.41 - 0.71) | 0.95 (0.90 - 0.99) |
| | Charges | Peri-Op (0: 243, 1: 137) | LR | 0.95 (0.91 - 0.98) | 0.87 (0.80 - 0.92) | 0.88 (0.82 - 0.93) | 0.97 (0.91 - 1.00) | 0.81 (0.72 - 0.90) | 0.96 (0.87 - 1.00) |
| | Average Pain | Intra-Op (0: 580, 1: 238) | XGBoost | 0.63 (0.54 - 0.70) | 0.71 (0.65 - 0.75) | 0.81 (0.76 - 0.84) | 0.76 (0.70 - 0.81) | 0.86 (0.81 - 0.91) | 0.33 (0.23 - 0.43) |

Abbreviations: AUC, area under the curve; CI, confidence interval; Intra-Op, intraoperative; LOS, length of hospital stay; LR, logistic regression; Peri-Op, perioperative; XGBoost, extreme gradient boosting.

**Table 3.** - Importance of Different Datasets

| Surgery Type | Outcome | Intra-Op | Peri-Op | Peri-Op Cognitive |
|---|---|---|---|---|
| **All surgeries** | LOS | | | ✓ |
| | Charges | | | ✓ |
| | Average Pain | | ✓ | |
| | 1-year Mortality | | | ✓ |
| **Orthopedics** | LOS | | | ✓ |
| | Charges | | | ✓ |
| | Average Pain | | ✓ | |
| **Neurology** | LOS | ✓ | | |
| | Charges | | | ✓ |
| | Average Pain | ✓ | | |
| **Cardiac and vascular** | LOS | | ✓ | |
| | Charges | | | ✓ |
| | Average Pain | ✓ | | |
| **Urology** | LOS | | | ✓ |
| | Charges | | | ✓ |
| | Average Pain | | | ✓ |
| **Gynecology** | LOS | | | ✓ |
| | Charges | ✓ | | |
| | Average Pain | | ✓ | |
| **Otolaryngology** | LOS | | | ✓ |
| | Charges | | ✓ | |

| | Average Pain | ✓ | | |
|---|---|---|---|---|

Abbreviations: Intra-Op, intraoperative; LOS, length of hospital stay; Peri-Op, perioperative.

**FIGURES**

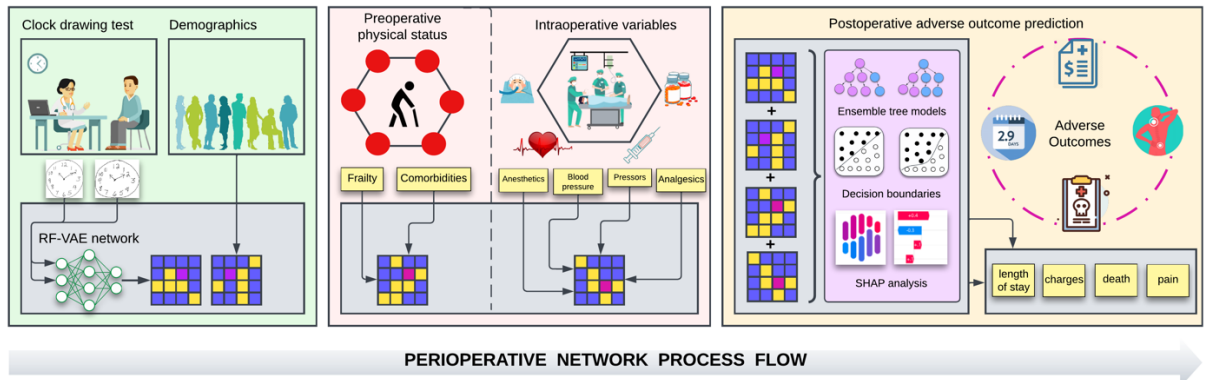

**Figure 1.:** Perioperative Network Process Flow. Cognitive screeners such as the clock drawing test and demographic variables are collected preoperatively. Frailty and comorbidities are ascertained before surgery. Anesthetic, analgesic, and hemodynamic variables are collected during the operation. All data are combined into ensemble-based models to ascertain the postoperative incidence of adverse outcomes such as longer length of stay, higher hospital charges, higher pain, and mortality. Abbreviations: RF-VAE, relevance-factor variational autoencoder; SHAP, Shapley Additive Explanations.

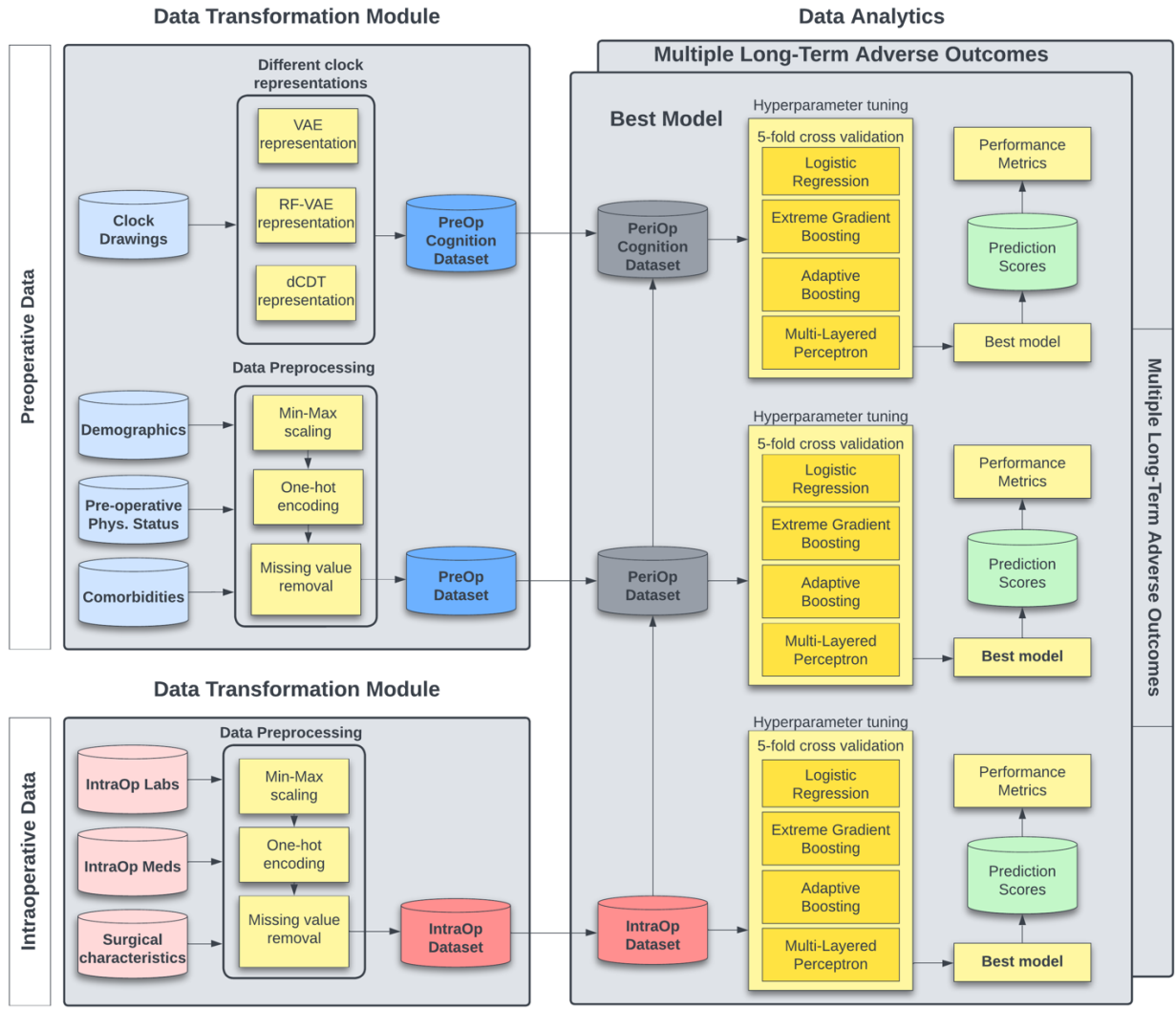

**Figure 2.:** Perioperative AI-Based Integrated Modeling of Surgeries (Peri-AIIMS) Framework. Intraoperative variables are converted into one-hot encoded vectors or min-max scaled numeric vectors. Cognitive status represented by clock drawings are projected onto the 10D latent RF-VAE latent space as described by Bandyopadhyay et al, 2023[25]. Demographics and comorbidities are one-hot encoded. Preoperative physical status variables (frailty and ASA) are min-max scaled. Multiple models are trained within 5-fold cross-validation using either the intraoperative, perioperative, or perioperative cognition datasets. In each case, the best model for each adverse

outcome is reported in the paper. Abbreviations: RF-VAE, relevance-factor variational autoencoder.

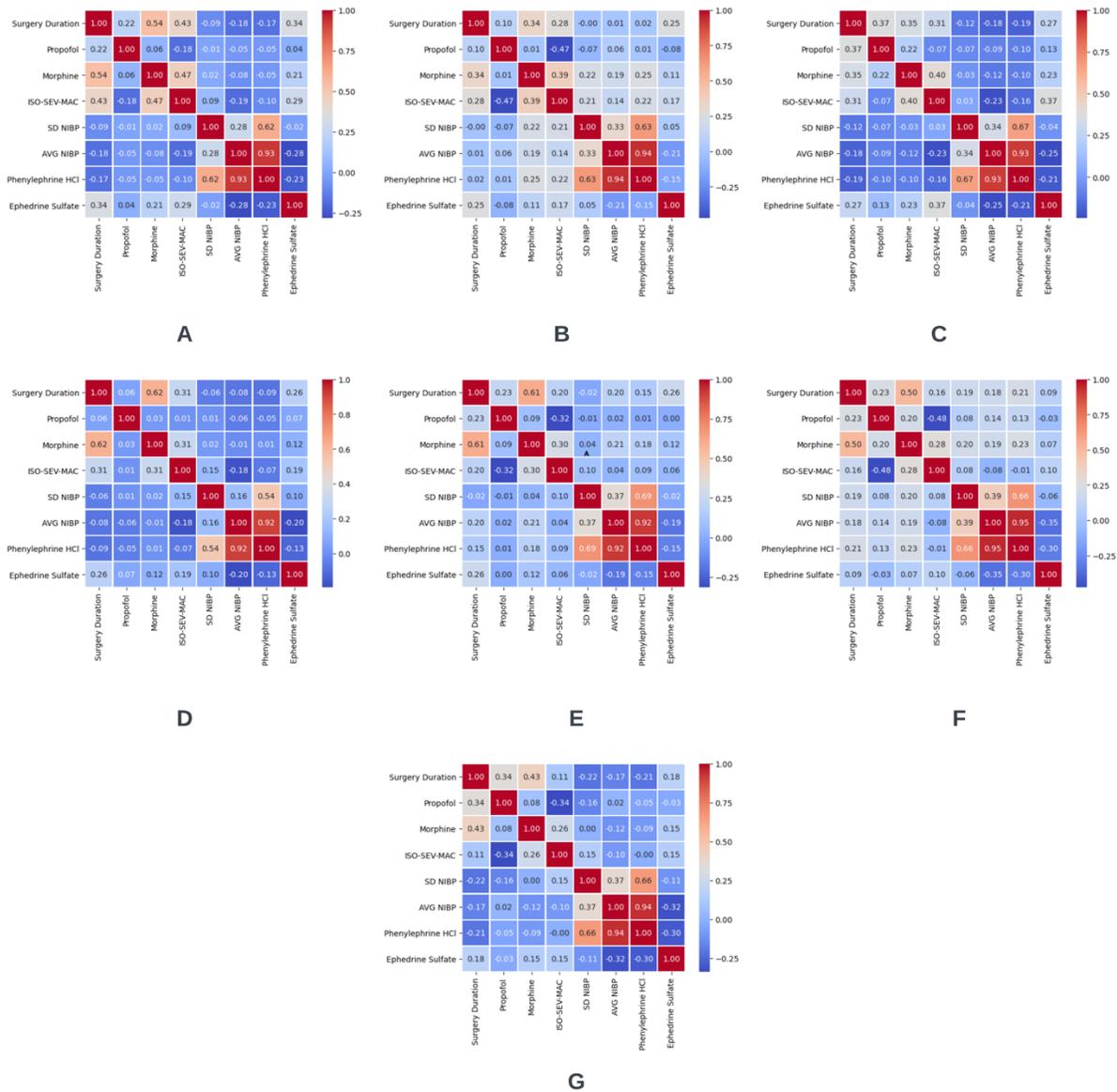

**Figure 3.:** Cross-Correlation Matrix Between the Different Intraoperative Features of Surgeries. (A) All surgeries, (B) Orthopedic surgeries, (C) Neurosurgeries, (D) Cardiovascular surgeries, (E) Urologic surgeries, (F) Gynecologic surgeries, and (G) Otolaryngologic surgeries. Abbreviations: AVG NIBP, average deviation of noninvasive

blood pressure; HCl, hydrochloride; iso-sev-MAC, mean alveolar concentration of isoflurane and sevoflurane; SD NIBP, standard deviation of noninvasive blood pressure.

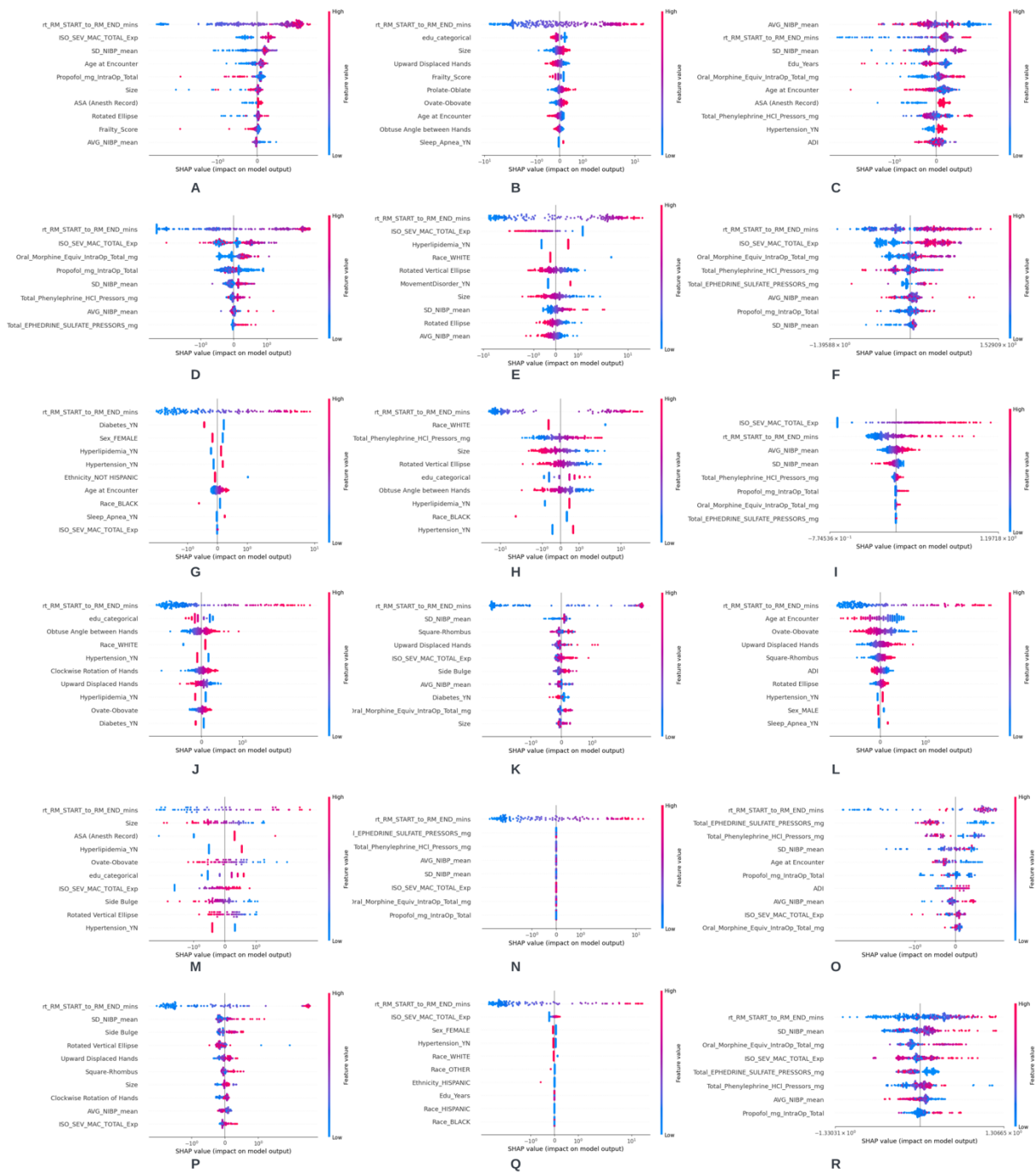

**Figure 4.:** SHAP Interpretation Plots for Best Models Showing Top 10 Features. (A) Orthopedic surgery LOS; (B) Orthopedic surgery charges; (C) Orthopedic surgery

average pain; (D) Neurosurgery LOS; (E) Neurosurgery charges; (F) Neurosurgery average pain; (G) Cardiac and vascular surgery LOS; (H) Cardiac and vascular surgery charges; (I) Cardiac and vascular surgery average pain; (J) Urologic surgery LOS; (K) Urologic surgery charges; (L) Urologic surgery average pain; (M) Gynecology surgery LOS; (N) Gynecology surgery charges; (O) Gynecology surgery average pain; (P) Otolaryngology surgery LOS; (Q) Otolaryngology surgery charges; (R) Otolaryngology surgery average pain. Abbreviations: LOS, length of hospital stay.

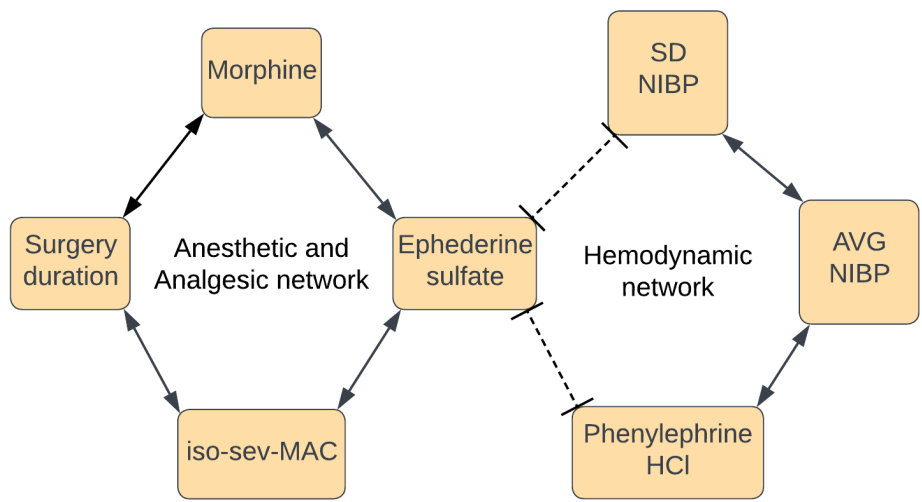

**Figure 5.:** Intraoperative Variables Correlation Network. This network is generated using the significant (according to relative Cohen's d values) positive and negative correlations between intraoperative variables. Dotted lines show negative correlation and solid lines show positive correlation. This shows the emergence of networks defining the 2 major housekeeping tasks during surgery: (A) anesthetic and analgesic management and (B) hemodynamic management. Disruptions within the positive correlations have led to adverse outcomes in different surgery-outcome combinations in the present study. Abbreviations: HCl, hydrochloride; iso-sev-mAC, mean alveolar

concentration of isoflurane and sevoflurane; SD NIBP, standard deviation of noninvasive blood pressure; AVG NIBP, average deviation of noninvasive blood pressure.

## SUPPLEMENTARY

**Supplemental Table 1.:** Classification Performance for All Surgeries Over Different Datasets

| Surgery Type | Outcome | Dataset | AUC (95% CI) | Accuracy (95% CI) | F1 Score (95% CI) | Precision (95% CI) | Sensitivity (95% CI) | Specificity (95% CI) |
|---|---|---|---|---|---|---|---|---|
| All Surgeries | LOS | Intra-Op | 0.92 (0.91 - 0.93) | 0.85 (0.84 - 0.86) | 0.81 (0.79 - 0.83) | 0.77 (0.74 - 0.79) | 0.86 (0.85 - 0.88) | 0.85 (0.83 - 0.86) |
| | | Peri-Op | 0.92 (0.91 - 0.93) | 0.83 (0.82 - 0.85) | 0.80 (0.78 - 0.81) | 0.74 (0.71 - 0.77) | 0.86 (0.84 - 0.88) | 0.82 (0.80 - 0.84) |
| | | Peri-Op cognitive | 0.93 (0.92 - 0.95) | 0.87 (0.85 - 0.89) | 0.83 (0.81 - 0.85) | 0.82 (0.79 - 0.84) | 0.85 (0.82 - 0.88) | 0.88 (0.86 - 0.90) |
| | Charges | Intra-Op | 0.98 (0.98 - 0.98) | 0.93 (0.92 - 0.94) | 0.93 (0.92 - 0.94) | 0.93 (0.92 - 0.94) | 0.93 (0.92 - 0.94) | 0.93 (0.92 - 0.94) |
| | | Peri-Op | 0.98 (0.98 - 0.98) | 0.92 (0.91 - 0.94) | 0.93 (0.92 - 0.94) | 0.94 (0.92 - 0.95) | 0.92 (0.90 - 0.93) | 0.93 (0.92 - 0.94) |
| | | Peri-Op cognitive | 0.98 (0.97 - 0.99) | 0.93 (0.92 - 0.94) | 0.93 (0.92 - 0.95) | 0.94 (0.93 - 0.96) | 0.92 (0.91 - 0.94) | 0.94 (0.92 - 0.95) |
| | Average Pain | Intra-Op | 0.83 (0.81 - 0.84) | 0.76 (0.75 - 0.77) | 0.79 (0.78 - 0.80) | 0.79 (0.78 - 0.81) | 0.78 (0.77 - 0.80) | 0.74 (0.71 - 0.76) |
| | | Peri-Op | 0.82 (0.80 - 0.84) | 0.77 (0.75 - 0.79) | 0.80 (0.78 - 0.82) | 0.80 (0.77 - 0.83) | 0.80 (0.77 - 0.82) | 0.73 (0.70 - 0.76) |
| | | Peri-Op cognitive | 0.82 (0.80 - 0.83) | 0.76 (0.74 - 0.78) | 0.79 (0.77 - 0.81) | 0.79 (0.77 - 0.82) | 0.79 (0.76 - 0.81) | 0.72 (0.69 - 0.75) |
| | Mortality | Intra-Op | 0.63 (0.59 - 0.68) | 0.64 (0.63 - 0.66) | 0.12 (0.10 - 0.15) | 0.07 (0.06 - 0.08) | 0.53 (0.45 - 0.61) | 0.64 (0.63 - 0.66) |
| | | Peri-Op | 0.70 (0.64 - 0.77) | 0.69 (0.67 - 0.71) | 0.15 (0.12 - 0.20) | 0.09 (0.06 - 0.11) | 0.64 (0.54 - 0.75) | 0.69 (0.67 - 0.71) |
| | | Peri-Op cognitive | 0.71 (0.66 - 0.77) | 0.70 (0.68 - 0.73) | 0.16 (0.11 - 0.20) | 0.09 (0.06 - 0.12) | 0.61 (0.50 - 0.73) | 0.71 (0.69 - 0.73) |

Abbreviations: AUC, area under the curve; CI, confidence interval; Intra-Op, intraoperative; LOS, length of hospital stay; Peri-Op, perioperative.

**Supplemental Table 2.:** Classification Performance for Orthopedic Surgeries Over Different Datasets

| Surgery Type | Outcome | Dataset | AUC (95% CI) | Accuracy (95% CI) | F1 Score (95% CI) | Precision (95% CI) | Sensitivity (95% CI) | Specificity (95% CI) |
|---|---|---|---|---|---|---|---|---|
| Orthopedic | LOS | Intra-Op | 0.82 (0.78 - 0.85) | 0.72 (0.69 - 0.76) | 0.80 (0.77 - 0.82) | 0.91 (0.89 - 0.94) | 0.71 (0.67 - 0.75) | 0.79 (0.73 - 0.84) |

| | | | AUC | Accuracy | F1 Score | Precision | Sensitivity | Specificity |
|---|---|---|---|---|---|---|---|---|
| | | Peri-Op | 0.83 (0.77 - 0.87) | 0.74 (0.70 - 0.78) | 0.81 (0.77 - 0.85) | 0.92 (0.87 - 0.95) | 0.72 (0.67 - 0.78) | 0.79 (0.68 - 0.87) |
| | | Peri-Op cognitive | 0.84 (0.79 - 0.88) | 0.83 (0.79 - 0.88) | 0.90 (0.87 - 0.93) | 0.85 (0.82 - 0.89) | 0.95 (0.92 - 0.97) | 0.47 (0.35 - 0.58) |
| | Charges | Intra-Op | 0.98 (0.96 - 1.00) | 0.96 (0.95 - 0.97) | 0.98 (0.97 - 0.99) | 1.00 (0.99 - 1.00) | 0.96 (0.94 - 0.97) | 0.99 (0.95 - 1.00) |
| | | Peri-Op | 0.99 (0.98 - 1.00) | 0.94 (0.92 - 0.96) | 0.97 (0.95 - 0.98) | 0.99 (0.98 - 1.00) | 0.94 (0.92 - 0.97) | 0.94 (0.82 - 1.00) |
| | | Peri-Op cognitive | 0.99 (0.99 - 1.00) | 0.95 (0.92 - 0.97) | 0.97 (0.96 - 0.98) | 1.00 (1.00 - 1.00) | 0.95 (0.91 - 0.97) | 1.00 (1.00 - 1.00) |
| | Average Pain | Intra-Op | 0.67 (0.62 - 0.71) | 0.83 (0.80 - 0.86) | 0.91 (0.89 - 0.92) | 0.85 (0.82 - 0.88) | 0.97 (0.96 - 0.98) | 0.18 (0.13 - 0.25) |
| | | Peri-Op | 0.69 (0.62 - 0.77) | 0.84 (0.80 - 0.88) | 0.91 (0.88 - 0.93) | 0.86 (0.82 - 0.89) | 0.97 (0.95 - 0.98) | 0.26 (0.16 - 0.38) |
| | | Peri-Op cognitive | 0.67 (0.58 - 0.75) | 0.82 (0.79 - 0.87) | 0.90 (0.88 - 0.93) | 0.84 (0.81 - 0.89) | 0.97 (0.95 - 0.99) | 0.13 (0.04 - 0.23) |

Abbreviations: AUC, area under the curve; CI, confidence interval; Intra-Op, intraoperative; LOS, length of hospital stay; Peri-Op, perioperative.

**Supplemental Table 3.:** Classification Performance for Neurosurgeries Over Different Datasets

| Surgery Type | Outcome | Dataset | AUC (95% CI) | Accuracy (95% CI) | F1 Score (95% CI) | Precision (95% CI) | Sensitivity (95% CI) | Specificity (95% CI) |
|---|---|---|---|---|---|---|---|---|
| Neurologic | LOS | Intra-Op | 0.92 (0.88 - 0.94) | 0.86 (0.82 - 0.90) | 0.89 (0.86 - 0.92) | 0.87 (0.83 - 0.91) | 0.91 (0.87 - 0.94) | 0.79 (0.72 - 0.84) |
| | | Peri-Op | 0.87 (0.82 - 0.92) | 0.79 (0.74 - 0.85) | 0.85 (0.81 - 0.90) | 0.80 (0.73 - 0.87) | 0.92 (0.86 - 0.97) | 0.55 (0.43 - 0.70) |
| | | Peri-Op cognitive | 0.82 (0.74 - 0.88) | 0.75 (0.68 - 0.82) | 0.80 (0.73 - 0.86) | 0.87 (0.79 - 0.92) | 0.73 (0.65 - 0.82) | 0.78 (0.65 - 0.89) |
| | Charges | Intra-Op | 0.91 (0.86 - 0.94) | 0.90 (0.88 - 0.92) | 0.94 (0.92 - 0.96) | 0.91 (0.88 - 0.94) | 0.98 (0.96 - 0.99) | 0.61 (0.51 - 0.70) |
| | | Peri-Op | 0.92 (0.87 - 0.96) | 0.90 (0.85 - 0.94) | 0.94 (0.90 - 0.97) | 0.93 (0.88 - 0.97) | 0.95 (0.90 - 0.98) | 0.68 (0.49 - 0.84) |
| | | Peri-Op cognitive | 0.93 (0.87 - 0.97) | 0.86 (0.80 - 0.91) | 0.91 (0.87 - 0.94) | 0.94 (0.89 - 0.98) | 0.88 (0.82 - 0.92) | 0.76 (0.60 - 0.91) |
| | Average Pain | Intra-Op | 0.74 (0.68 - 0.80) | 0.78 (0.74 - 0.82) | 0.86 (0.83 - 0.88) | 0.81 (0.77 - 0.86) | 0.90 (0.87 - 0.93) | 0.49 (0.41 - 0.58) |
| | | Peri-Op | 0.71 (0.62 - 0.81) | 0.71 (0.63 - 0.77) | 0.79 (0.71 - 0.83) | 0.86 (0.77 - 0.93) | 0.72 (0.63 - 0.80) | 0.67 (0.52 - 0.82) |
| | | Peri-Op cognitive | 0.63 (0.53 - 0.74) | 0.73 (0.66 - 0.79) | 0.83 (0.77 - 0.87) | 0.77 (0.69 - 0.83) | 0.89 (0.84 - 0.94) | 0.30 (0.18 - 0.45) |

Abbreviations: AUC, area under the curve; CI, confidence interval; Intra-Op, intraoperative; LOS, length of hospital stay; Neuro, neurosurgery; Peri-Op, perioperative.

**Supplemental Table 4.:** Classification Performance for Cardiovascular and Vascular Surgeries Over Different Datasets

| Surgery Type | Outcome | Dataset | AUC (95% CI) | Accuracy (95% CI) | F1 Score (95% CI) | Precision (95% CI) | Sensitivity (95% CI) | Specificity (95% CI) |
|---|---|---|---|---|---|---|---|---|
| Cardiovascular | LOS | Intra-Op | 0.87 (0.84 - 0.91) | 0.78 (0.73 - 0.82) | 0.79 (0.74 - 0.83) | 0.88 (0.81 - 0.93) | 0.71 (0.64 - 0.78) | 0.87 (0.81 - 0.92) |
| | | Peri-Op | 0.89 (0.83 - 0.93) | 0.83 (0.78 - 0.88) | 0.85 (0.80 - 0.90) | 0.86 (0.79 - 0.94) | 0.84 (0.76 - 0.92) | 0.82 (0.73 - 0.91) |
| | | Peri-Op cognitive | 0.87 (0.82 - 0.92) | 0.81 (0.75 - 0.86) | 0.83 (0.76 - 0.88) | 0.89 (0.80 - 0.94) | 0.78 (0.69 - 0.86) | 0.86 (0.75 - 0.92) |
| | Charges | Intra-Op | 0.95 (0.93 - 0.97) | 0.86 (0.83 - 0.90) | 0.91 (0.88 - 0.93) | 1.00 (1.00 - 1.00) | 0.83 (0.79 - 0.87) | 1.00 (1.00 - 1.00) |
| | | Peri-Op | 0.96 (0.93 - 0.98) | 0.91 (0.85 - 0.95) | 0.94 (0.90 - 0.97) | 0.98 (0.95 - 1.00) | 0.90 (0.82 - 0.95) | 0.95 (0.82 - 1.00) |
| | | Peri-Op cognitive | 0.96 (0.93 - 0.99) | 0.92 (0.87 - 0.95) | 0.94 (0.91 - 0.97) | 0.98 (0.96 - 1.00) | 0.91 (0.86 - 0.95) | 0.93 (0.86 - 1.00) |
| | Average Pain | Intra-Op | 0.70 (0.64 - 0.76) | 0.64 (0.59 - 0.70) | 0.69 (0.64 - 0.75) | 0.76 (0.70 - 0.83) | 0.63 (0.55 - 0.70) | 0.66 (0.56 - 0.74) |
| | | Peri-Op | 0.66 (0.58 - 0.76) | 0.60 (0.53 - 0.68) | 0.66 (0.59 - 0.74) | 0.73 (0.63 - 0.84) | 0.62 (0.52 - 0.71) | 0.58 (0.43 - 0.70) |
| | | Peri-Op cognitive | 0.66 (0.56 - 0.75) | 0.61 (0.54 - 0.68) | 0.66 (0.58 - 0.73) | 0.76 (0.65 - 0.83) | 0.60 (0.51 - 0.69) | 0.63 (0.49 - 0.75) |

Abbreviations: AUC, area under the curve; CI, confidence interval; Intra-Op, intraoperative; LOS, length of hospital stay; Peri-Op, perioperative.

**Supplemental Table 5.:** Classification Performance for Urologic Surgeries Over Different Datasets

| Surgery Type | Outcome | Dataset | AUC (95% CI) | Accuracy (95% CI) | F1 Score (95% CI) | Precision (95% CI) | Sensitivity (95% CI) | Specificity (95% CI) |
|---|---|---|---|---|---|---|---|---|
| Urologic | LOS | Intra-Op | 0.71 (0.66 - 0.76) | 0.74 (0.71 - 0.78) | 0.53 (0.47 - 0.58) | 0.50 (0.42 - 0.58) | 0.56 (0.48 - 0.64) | 0.81 (0.77 - 0.85) |
| | | Peri-Op | 0.71 (0.62 - 0.78) | 0.77 (0.71 - 0.81) | 0.47 (0.37 - 0.58) | 0.61 (0.46 - 0.73) | 0.39 (0.28 - 0.50) | 0.91 (0.86 - 0.93) |

| | | | AUC (95% CI) | Accuracy (95% CI) | F1 Score (95% CI) | Precision (95% CI) | Sensitivity (95% CI) | Specificity (95% CI) |
|---|---|---|---|---|---|---|---|---|
| | | Peri-Op cognitive | 0.80 (0.71 - 0.87) | 0.77 (0.71 - 0.82) | 0.63 (0.52 - 0.72) | 0.57 (0.44 - 0.69) | 0.70 (0.59 - 0.81) | 0.80 (0.74 - 0.86) |
| | Charges | Intra-Op | 0.97 (0.95 - 0.98) | 0.91 (0.88 - 0.93) | 0.88 (0.85 - 0.91) | 0.98 (0.95 - 0.99) | 0.80 (0.75 - 0.86) | 0.98 (0.97 - 1.00) |
| | | Peri-Op | 0.95 (0.92 - 0.99) | 0.92 (0.89 - 0.96) | 0.91 (0.87 - 0.95) | 0.94 (0.90 - 0.99) | 0.88 (0.89 - 0.94) | 0.96 (0.93 - 0.99) |
| | | Peri-Op cognitive | 0.96 (0.93 - 0.99) | 0.93 (0.89 - 0.96) | 0.91 (0.86 - 0.95) | 0.94 (0.88 − 1.00) | 0.88 (0.81 - 0.94) | 0.96 (0.92 − 1.00) |
| | Average Pain | Intra-Op | 0.73 (0.67 - 0.77) | 0.67 (0.62 - 0.70) | 0.64 (0.59 - 0.68) | 0.84 (0.78 - 0.89) | 0.52 (0.46 - 0.58) | 0.86 (0.81 - 0.91) |
| | | Peri-Op | 0.75 (0.69 - 0.81) | 0.69 (0.62 - 0.75) | 0.70 (0.62 - 0.77) | 0.76 (0.68 - 0.85) | 0.65 (0.53 - 0.74) | 0.74 (0.64 - 0.83) |
| | | Peri-Op cognitive | 0.75 (0.69 - 0.81) | 0.70 (0.64 - 0.76) | 0.70 (0.64 - 0.76) | 0.78 (0.68 - 0.86) | 0.64 (0.55 - 0.72) | 0.78 (0.69 - 0.88) |

Abbreviations: AUC, area under the curve; CI, confidence interval; Intra-Op, intraoperative; LOS, length of hospital stay; Peri-Op, perioperative.

**Supplemental Table 6.:** Classification Performance for Gynecologic Surgeries Over Different Datasets

| Surgery Type | Outcome | Dataset | AUC (95% CI) | Accuracy (95% CI) | F1 Score (95% CI) | Precision (95% CI) | Sensitivity (95% CI) | Specificity (95% CI) |
|---|---|---|---|---|---|---|---|---|
| Gynecologic | LOS | Intra-Op | 0.78 (0.69 - 0.86) | 0.67 (0.57 - 0.74) | 0.60 (0.46 - 0.70) | 0.51 (0.36 - 0.61) | 0.71 (0.60 - 0.86) | 0.63 (0.52 - 0.74) |
| | | Peri-Op | 0.76 (0.64 - 0.88) | 0.68 (0.57 - 0.79) | 0.52 (0.33 - 0.67) | 0.44 (0.27 - 0.64) | 0.65 (0.40 - 0.83) | 0.71 (0.56 - 0.82) |
| | | Peri-Op cognitive | 0.80 (0.61 - 0.91) | 0.72 (0.60 - 0.84) | 0.56 (0.28 - 0.75) | 0.53 (0.23 - 0.77) | 0.60 (0.31 - 0.82) | 0.79 (0.66 - 0.89) |
| | Charges | Intra-Op | 0.99 (0.97 - 1.00) | 0.94 (0.90 - 0.97) | 0.96 (0.92 - 0.98) | 1.00 (1.00 - 1.00) | 0.91 (0.84 - 0.95) | 1.00 (1.00 - 1.00) |
| | | Peri-Op | 0.98 (0.94 - 1.00) | 0.94 (0.85 - 0.98) | 0.95 (0.88 - 0.99) | 0.95 (0.88 - 1.00) | 0.94 (0.87 - 1.00) | 0.91 (0.76 - 1.00) |
| | | Peri-Op cognitive | 0.97 (0.93 - 0.99) | 0.86 (0.79 - 0.95) | 0.89 (0.82 - 0.96) | 0.97 (0.91 - 1.00) | 0.82 (0.72 - 0.92) | 0.95 (0.85 - 1.00) |
| | Average Pain | Intra-Op | 0.65 (0.48 - 0.82) | 0.86 (0.79 - 0.91) | 0.92 (0.88 - 0.95) | 0.88 (0.82 - 0.94) | 0.96 (0.93 - 0.99) | 0.24 (0.07 - 0.48) |
| | | Peri-Op | 0.75 (0.60 - 0.87) | 0.80 (0.70 - 0.86) | 0.88 (0.81 - 0.92) | 0.88 (0.77 - 0.94) | 0.88 (0.80 - 0.96) | 0.36 (0.11 - 0.65) |
| | | Peri-Op cognitive | 0.61 (0.47 - 0.76) | 0.72 (0.62 - 0.81) | 0.83 (0.76 - 0.89) | 0.82 (0.71 - 0.91) | 0.85 (0.75 - 0.93) | 0.10 (0.00 - 0.27) |

Abbreviations: AUC, area under the curve; CI, confidence interval; Intra-Op,

intraoperative; LOS, length of hospital stay; Peri-Op, perioperative.

**Supplemental Table 7.:** Classification Performance for Otolaryngologic Surgeries Over

Different Datasets

| Surgery Type | Outcome | Dataset | AUC (95% CI) | Accuracy (95% CI) | F1 Score (95% CI) | Precision (95% CI) | Sensitivity (95% CI) | Specificity (95% CI) |
|---|---|---|---|---|---|---|---|---|
| Otolaryngologic | LOS | Intra-Op | 0.85 (0.80 - 0.90) | 0.85 (0.81 - 0.88) | 0.69 (0.57 - 0.75) | 0.84 (0.73 - 0.92) | 0.58 (0.47 - 0.66) | 0.96 (0.93 - 0.98) |
| | | Peri-Op | 0.88 (0.81 - 0.93) | 0.79 (0.73 - 0.86) | 0.65 (0.53 - 0.74) | 0.71 (0.56 - 0.82) | 0.59 (0.49 - 0.74) | 0.89 (0.82 - 0.94) |
| | | Peri-Op cognitive | 0.88 (0.80 - 0.94) | 0.82 (0.75 - 0.88) | 0.69 (0.52 - 0.78) | 0.86 (0.71 - 0.95) | 0.58 (0.41 - 0.71) | 0.95 (0.90 - 0.99) |
| | Charges | Intra-Op | 0.95 (0.92 - 0.97) | 0.85 (0.81 - 0.90) | 0.87 (0.83 - 0.91) | 0.91 (0.85 - 0.94) | 0.85 (0.78 - 0.90) | 0.87 (0.81 - 0.92) |
| | | Peri-Op | 0.95 (0.91 - 0.98) | 0.87 (0.80 - 0.92) | 0.88 (0.82 - 0.93) | 0.97 (0.91 - 1.00) | 0.81 (0.72 - 0.90) | 0.96 (0.87 - 1.00) |
| | | Peri-Op cognitive | 0.90 (0.83 - 0.94) | 0.84 (0.78 - 0.88) | 0.88 (0.83 - 0.92) | 0.86 (0.79 - 0.92) | 0.91 (0.83 - 0.96) | 0.70 (0.52 - 0.83) |
| | Average Pain | Intra-Op | 0.63 (0.54 - 0.70) | 0.71 (0.65 - 0.75) | 0.81 (0.76 - 0.84) | 0.76 (0.70 - 0.81) | 0.86 (0.81 - 0.91) | 0.33 (0.23 - 0.43) |
| | | Peri-Op | 0.60 (0.46 - 0.73) | 0.72 (0.64 - 0.79) | 0.82 (0.76 - 0.87) | 0.78 (0.70 - 0.84) | 0.87 (0.80 - 0.93) | 0.31 (0.16 - 0.44) |
| | | Peri-Op cognitive | 0.58 (0.45 - 0.70) | 0.68 (0.61 - 0.75) | 0.80 (0.75 - 0.85) | 0.73 (0.66 - 0.81) | 0.88 (0.82 - 0.94) | 0.06 (0.00 - 0.19) |

Abbreviations: AUC, area under the curve; CI, confidence interval; Intra-Op,

intraoperative; LOS, length of hospital stay; Peri-Op, perioperative.

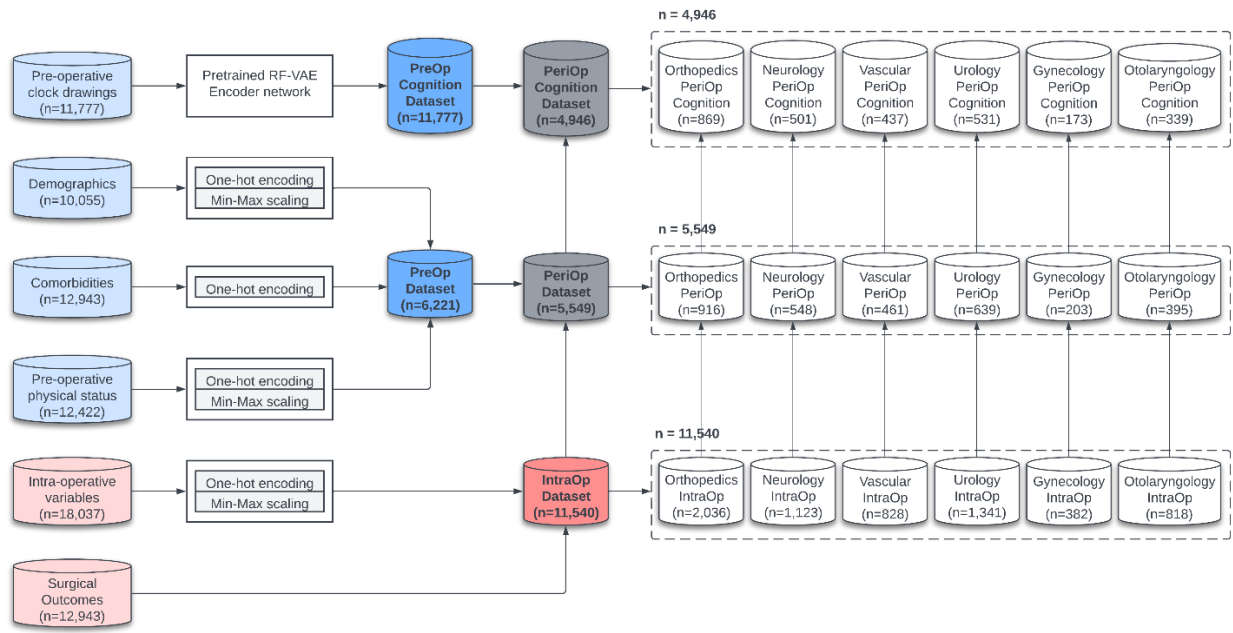

**Supplemental Figure 1.:** Data Modalities Consensus Diagram. This figure illustrates the stepwise inclusion of different data modalities to create the *intraoperative, perioperative,* and *perioperative cognitive datasets* used in this study. Preoperative clock drawings are projected onto a 10-dimensional RF-VAE DL network encoder. Categorical variables present in demographics, comorbidities, physical status, and intraoperative features are one-hot encoded, while numerical variables in these datasets are min-max scaled. Abbreviations: DL, deep learning; RF-VAE, relevance-factor variational autoencoder.

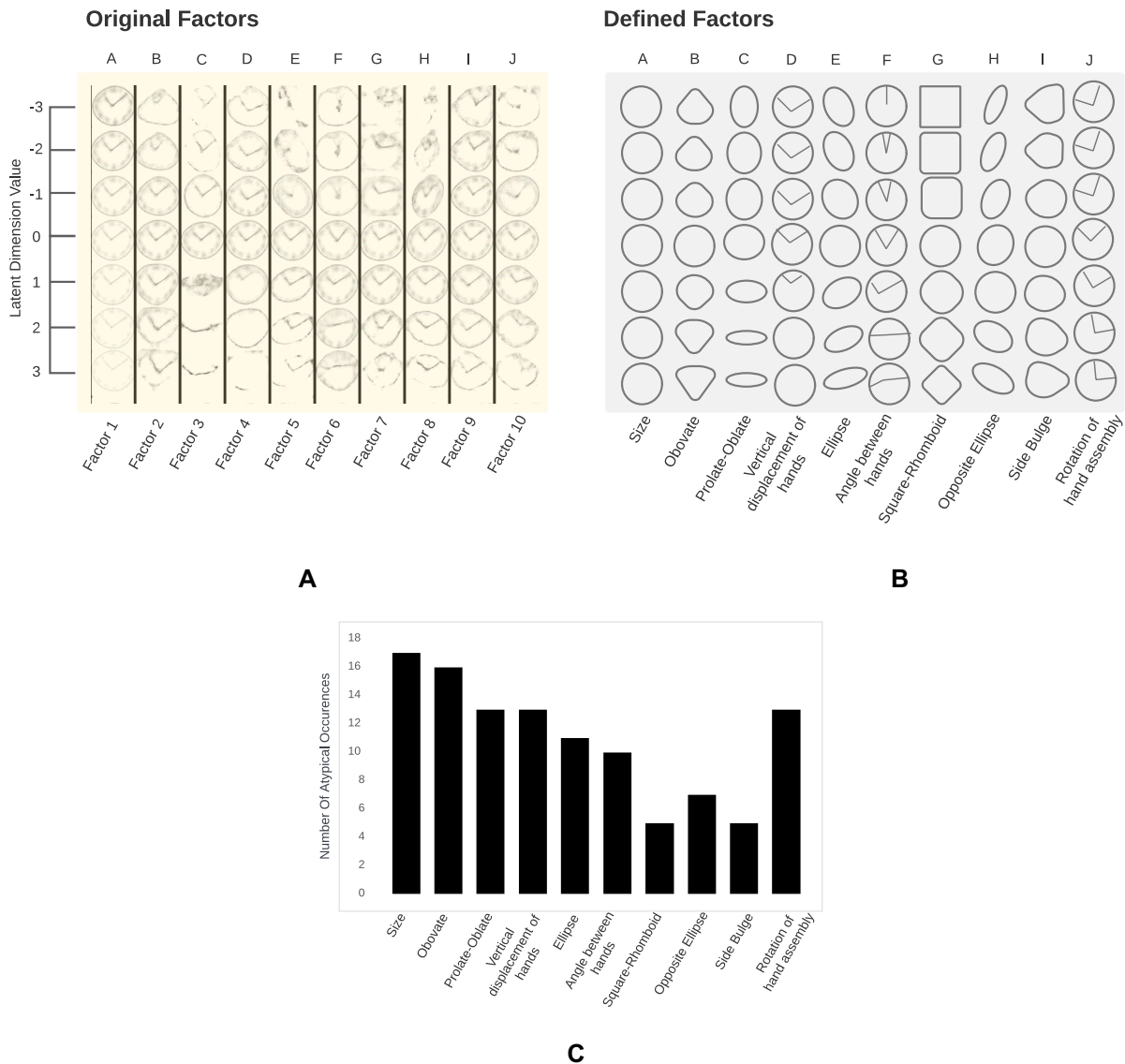

**Supplemental Figure 2.:** Depiction of the 10 Unique Constructional Features of Clock Drawings Discovered by the RF-VAE Latent Space. (A) RF-VAE latent representation of mutually disentangled factors; (B) Explanation of the factors; (C) Frequency of atypical occurrence of the factors in dementia patients compared to non-dementia peers. These figures were adapted from Bandyopadhyay et al 2023[25]. Abbreviations: RF-VAE, relevance-factor variational autoencoder.